\documentclass[11pt]{article}
\usepackage{acl2014}
\usepackage{times}
\usepackage{latexsym}

\usepackage{subcaption}
\usepackage{forest}
\usepackage{adjustbox}
\usepackage{array}
\usepackage{algorithm}
\usepackage{algpseudocode}
\usepackage{widetext}
\usepackage{url}

\title{Any-gram Kernels for Sentence Classification: \\ A Sentiment Analysis Case Study}

\author{Rasoul Kaljahi \\
		ADAPT Centre \\
		School of Computing \\
		Dublin City University, Ireland \\
		{\tt rasoul.kaljahi@adaptcentre.ie} \\\And
		Jennifer Foster \\
		ADAPT Centre \\
		School of Computing \\
		Dublin City University, Ireland \\
		{\tt jfoster@computing.dcu.ie} \\}

\date{}

\begin{document}
	
\maketitle

\begin{abstract}
	Any-gram kernels are a flexible and efficient way to employ bag-of-n-gram features when learning from textual data. They are also compatible with the use of word embeddings so that word similarities can be accounted for. While the original any-gram kernels are implemented on top of tree kernels, we propose a new approach which is independent of tree kernels and is more efficient. We also propose a more effective way to make use of 
	word embeddings than the original any-gram formulation. When applied to the task of sentiment classification, 
	our new formulation achieves significantly better performance.
\end{abstract}

\section{Introduction}\label{sec:intro}

The most basic and readily available features for machine learning of natural language are the words themselves.
Despite their simplicity, words have proven to be an essential part of statistical text analysis \cite{Pang-EtAl-EMNLP-2002,Joachims-ECML-1998}.
Words have often been presented to the learning algorithms in the form of \textit{bag-of-n-grams}, where $n$ (\textit{order of n-gram}) specifies the number of consecutive words grouped together.

There are, however, two main disadvantages to using bag-of-n-gram features.
Firstly, the optimum value of $n$ seems to be dependent on the problem and the data in hand.
For example, for the task of sentiment polarity classification, while some work report favourable results for bag-of-unigrams ($n=1$), others find higher orders to be more useful \cite{Pang-Lee-2008}.
Secondly, bag-of-n-grams suffer from high-dimensionality and sparsity, which is exacerbated as $n$ increases.

\textit{Kernel methods} \cite{Cristianini-Shawe-Taylor-Cambridge-2000} for machine learning can provide a way to alleviate the latter problem, as they enable the modelling of an exponential number of features implicitly without the need to extract explicit features.
In principle, kernel functions measure the similarity between every pair of instances in the data set.
These measures then become the new feature space on which the learning algorithm operates.

\textit{Tree kernels} are an example of kernel functions which have been used along with kernel-based machine learning algorithms such as support vector machines \cite{Moschitti-EtAl-2006} and perceptron \cite{Collins-Duffy-2002}.
Applied to tree structures usually generated by syntactic or semantic analyzers, tree kernels measure the similarity of these structures based on the number  of common tree fragments between them.
To exploit this feature of tree kernels in tackling the problems associated with learning from bag-of-n-gram features, in our previous work \cite{Kaljahi-Foster-PEOPLES-2016}, we proposed \textit{any-gram kernels}, which formats the input sentences as binary trees and uses tree kernels to compute the kernel.
This approach models bag-of-n-gram features where all possible orders of n-grams are automatically included in the feature space, accounting both for the problem of finding the optimum $n$ and for the problem of the high-dimensional sparse feature space.

\begin{figure}
	\footnotesize
	\begin{forest}
		[root [Good [X]][Good [{,} [X]][{,} [fast [X]][fast [service [X]][service [. [X]][.]]]]]]
	\end{forest}
	\caption{Representation of sentence \textit{Good, fast service.} for tree-kernel-based any-gram kernels}
	\label{fig:ngtk}
\end{figure}
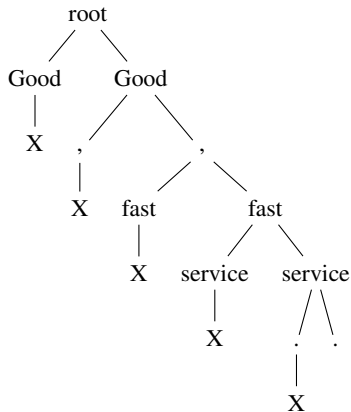

In order to represent a sentence as a binary tree in such a way that the tree kernel can extract any possible n-gram in the sentence as a tree fragment, every node is duplicated and additional nodes are required. An example is shown in Figure \ref{fig:ngtk} for the sentence \textit{Good, fast service.} 
The $X$ nodes make the extraction of unigrams possible, because the smallest tree fragment extracted by tree kernels is a production rule rather than a node.
This overhead is caused by the constraint imposed by the rather artificial use of tree kernels.
In this work, we propose an algorithm, independent of tree kernels, which eliminates the requirement that the input sentence be represented as a binary tree by instead treating the sentence as an array of tokens.

In the previous work \cite{Kaljahi-Foster-PEOPLES-2016}, we also modified the tree function so that the any-gram kernels can exploit the similarity between words obtained by comparing their pre-trained embedding vectors. 
We however found that these similarities in the any-gram kernels do not lead to consistently better results than the use of simple string match.
We propose a new method that can better leverage word embedding similarities in the context of any-gram kernels, yielding a higher performance.

We compare the performance of our any-gram approach to deep recurrent neural networks in the 
form of uni- and bidirectional LSTMs \cite{Hochreiter-Schmidhuber-NC-1997} on two widely used sentiment polarity classification data sets -- the SemEval 2014 aspect-based sentiment analysis data set \cite{Pontiki-EtAl-SemEval-2014} and the Stanford Sentiment Treebank \cite{Socher-EtAl-EMNLP-2013} -- and find that the any-gram performance is competitive with the LSTM performance. One advantage of the any-gram approach over the neural approach is that there is only one parameter to be tuned.

In the rest of this paper, we first review related work on kernel methods in Section \ref{sec:rw}. We then explain our new implementation of any-gram kernels in Section \ref{sec:anygram} and Section \ref{sec:anygram-we}.
In Section \ref{sec:expr}, the experiments are presented and the results discussed.
Finally, the paper is concluded in Section \ref{sec:conclusion} with a summary of the findings and some suggestions
for further work in this area.

\section{Related Work}\label{sec:rw}

We introduced any-gram kernels in our previous work \cite{Kaljahi-Foster-PEOPLES-2016} in an attempt to address the sparsity of bag-of-n-gram features as well as to reduce the efforts in engineering these features in terms of finding the most useful order of n-gram, i.e. $n$.
Those experiments show that any-grams can effectively replace bag-of-n-gram features without compromising performance.

As well as widely used general-purpose kernel functions such as \textit{polynomial} kernels and \textit{Radial Basis Function} (RBF) kernels, tree kernels \cite{Collins-Duffy-2002,Moschitti-2006} have been one of the most popular kernel functions applied to various natural language processing tasks such as parsing and tagging \cite{Collins-Duffy-2002}, semantic role labelling \cite{Moschitti-EtAl-2006}, quality estimation \cite{Hardmeier-EtAl-2012} and sentiment analysis \cite{Tu-EtAl-2012,Kaljahi-Foster-PEOPLES-2016}.
These kernels have been mainly used to learn problems in which knowledge of sentencial syntactic and/or semantic structure is useful. The input to the kernels are often constituency or dependency
trees produced by syntactic and/or semantic parsers.

\newcite{Srivastava-EtAl-EMNLP-2013} extend tree kernels to graph kernels, based on the work of \newcite{Gartner-EtAl-LNCS-2003} on random-walk-based graph kernels.
Treating dependency trees as graphs, the kernel value is computed by summation over the similarities of pairs of equal-length random walks on the tree pairs.
The similarities of these walks are computed as the product of pairwise similarities of nodes within the walks.

\textit{String kernels} \cite{Lodhi-EtAl-JMLR-2002}, also known as \textit{sequence kernels}, have also been used for text classification.
Originally, string kernels were applied at the character level, measuring the similarity between document pairs in terms of the number of matching subsequences in them.
Although these kernels allow non-contiguous subsequences, they work on fixed-length subsequences.
\newcite{Cancedda-EtAl-JMLR-2003} extend the character-level string kernels to the word level, as in the present work, arguing that this level of granularity is more efficient and also linguistically motivated.
Similar to string kernels, \textit{word-sequence} kernels also require the sequence length to be specified.
However, \newcite{Cancedda-EtAl-JMLR-2003} suggest that using dynamic programming techniques allows all subsequences of up to length $n$ to be considered with only a marginal increase in computational time.
Unlike string and word-sequence kernels, the length of the sequence does not have to be specified in advance for the any-gram kernels, making them more flexible.

The idea of replacing rigid string match with soft similarities in the computation of various kernel functions has also been investigated by some other scholars.
\newcite{Siolas-dAlche-Buc-IEEE-INNS-ENNS-2000} propose a document similarity metric for document categorization, which is based on the semantic proximity of word pairs defined as the inverse of the distance between them in the hypernym hierarchy of WordNet \cite{Miller-EtAl-IJL-1990}.
This metric can be incorporated into the RBF kernel of support vector machines or used in a K-nearest neighbour algorithm.
\newcite{Bloehdorn-ICDM-2006} extend this semantic similarity metric by capitalizing on the tree structure of  WordNet and including the depth of the concepts in the tree as a complement to the distance measure.

\newcite{Kim-EtAl-EMNLP-2015} use a polynomial kernel over the cosine similarity of word embedding vectors to compute \textit{word kernels} for short text categorization tasks including sentiment classification, where the kernel function is applied to a pair of words.
They extend the model to \textit{phrase} and \textit{sentence kernels} by aggregating the word kernels over the words forming the phrases and sentences.
These models also require the phrase length to be fixed.

The graph kernel method of \newcite{Srivastava-EtAl-EMNLP-2013} allows the node similarity based on the word embedding vectors of their corresponding words to be computed.
They use SENNA word embeddings \cite{Collobert-EtAl-JMLR-2011} and apply the kernel to sentiment classification, paraphrase detection and metaphor identification.

\newcite{Cancedda-EtAl-JMLR-2003} also propose that their word-sequence kernel can be extended to rectify the shortcoming of exact symbol matching by using a document-term matrix which estimates term-term similarity, as in information retrieval techniques.

\section{Any-gram Kernels}\label{sec:anygram}

The original implementation of any-gram kernels is built on top of tree kernels and requires the input text to be represented as a binary tree in a special format which contains $3N+1$ nodes for a sentence of length $N$.
In this section, we present an algorithm which does not require the input text to be represented as a tree, avoiding the overhead, while using the same principle to compute the kernel values.
This algorithm treats an input sentence as an array of tokens and uses dynamic programming to compute the kernel value for every pair of sentence instances.
The kernel function is defined as:

\begin{equation}\label{eq:any-gram}
K(S_1,S_2) = \sum_{i=1}^{len(S_1)}\sum_{j=1}^{len(S_2)}\Delta(S_1^i,S_2^j)
\end{equation}

where $\Delta(S_1^i,S_2^j)$ is the similarity between the $i^{th}$ token of sentence $S_1$ and $j^{th}$ token of sentence $S_2$, which is calculated as:

\begin{equation}\label{eq:agk:delta}
\small
\Delta(S_1^i,S_2^j) = 
\left\{\def\arraystretch{1.5}
\begin{array}{ll}
0~:~if~S_1^i \neq S_2^j~~
\vspace{.5em}
\\
\lambda \times (1+\Delta(S_1^{i+1},S_2^{j+1}))~:~if~S_1^i = S_2^j
\end{array}
\right.
\end{equation}

where $\lambda$ is the decay factor which penalizes the contribution of longer n-grams which get a bigger $\Delta$ value, and $S_1^{i+1}$ and $S_2^{j+1}$ are the tokens following the current tokens, $S_1^{i}$ and $S_2^{j}$, in the sentence.

Algorithm \ref{algo:agk-sm} shows how these definitions are implemented.\footnote{The implementation of the any-gram kernels is publicly available at \url{https://opengogs.adaptcentre.ie/rszk/agk}.}
The outer loop of the algorithm scans the tokens of the first sentence ($S_1$) in reverse order, and the inner loop scans the tokens of the second sentence ($S_2$) in the original order.
At each iteration, a cell in the $delta$ table, which corresponds to the $\Delta$ in Equation \ref{eq:agk:delta}, is filled with a value which is calculated based on the $\Delta$ value for the tokens following the current tokens in their respective sentences.
The latter value is already calculated and present in the table thanks to the reverse iteration of the outer loop.
The \textit{if} statement in the inner loop controls the execution of the statement computing the $\Delta$: if the current tokens are not equal, the computation is ignored.
Similar to the efficient implementation of the tree kernel algorithm\footnote{The tree kernel implementation initially collects all pairs of equal nodes in the instance pair and then performs the process over these nodes.} (and hence the original any-gram kernels), this is another key to the efficiency of the algorithm since in practice the majority of the token pairs in a sentence pair are not equal\footnote{Even though the computational complexity of the new any-gram kernels remains the same as the original ones, the running time of the algorithm is reduced due to the elimination of repetition and addition of nodes.}.
The kernel value for the input sentence pair is finally calculated by summing all the cells in the $delta$ table.

\begin{algorithm*}[t]
	\caption{: Any-gram kernel algorithm based on string match}\label{algo:agk-sm}
	\begin{algorithmic}
		\Function{anygram\_kernel\_sm}{$S_1, S_2$}
		\Comment{Inputs: tokenized sentence 1 and sentence 2}
		\State $kernel \gets 0$
		\State $delta[1, ..., len(S_1) + 1][1, ..., len(S_2) + 1] \gets 0$
		\Comment{Initializing $\Delta$ table}
		\State $L_1 \gets len(S_1)$
		\State $L_2 \gets len(S_2)$
		\For{$i \in \{L_1, ..., 1\}$}
		\Comment{Looping through sentence 1}
		\For{$j \in \{1, ..., L_2\}$}
		\Comment{Looping through sentence 2}
		\If{$S_1[i] = S_2[j]$}
		\Comment{Compute $\Delta$ only if tokens are equal}
		\State $delta[i][j] \gets \lambda \times (1 + delta[i+1][j+1])$
		\State $kernel \gets kernel + delta[i][j]$
		\EndIf
		\EndFor
		\EndFor
		\State \Return $kernel$
		\EndFunction
	\end{algorithmic}
\end{algorithm*}

\section{Word Embeddings in Any-gram Kernels}\label{sec:anygram-we}

In order to exploit the semantic similarity between words and lift the constraint of exact similarity of words (string match) in computing the kernel, we extended this model in \cite{Kaljahi-Foster-PEOPLES-2016} to incorporate word embeddings in the kernel computation.
The model is based on a \textit{similarity threshold} value which considers two nodes similar if the (cosine) similarity of their corresponding embedding vectors passes a certain threshold.
In other words, two nodes are considered the same if their embedding vectors are at least as similar as the threshold specifies.
To this end, the original tree kernel function is modified as follows:

\vspace{-1em}
\begin{widetext}
	\begin{equation}\label{eq:tk:delta2}
	\Delta(n_1,n_2) = 
	\left\{\def\arraystretch{1.5}
	\begin{array}{ll}
	0~:~if~|pr_1| != |pr_2|~~or~~\prod_{i=1}^{|pr_1|}\Theta(n_{(pr_1)i}, n_{(pr_2)i}) = 0 
	\vspace{.5em}
	\\
	1~:~if~n_1,~n_2~are~pre\mbox{-}terminals~~and~~\prod_{i=1}^{|pr_1|}\Theta(n_{(pr_1)i}, n_{(pr_2)i}) = 1 
	\vspace{.5em}
	\\
	\prod_{j=1}^{nc}(1+\Delta(c_{n_1}^j,c_{n_2}^j)): otherwise
	\end{array}
	\right.
	\end{equation}
\end{widetext}

where $pr_1$ and $pr_2$ are the production rules rewriting $n_1$ and $n_2$ respectively, $|pr_1|$ and $|pr_2|$ are the number of nodes in the production rules, $n_{(pr_1)i}$ and $n_{(pr_2)i}$ are the $i^{th}$ peer nodes of the two production rules and $nc$ is the number of children of $n_1$ (and $n_2$ as they are equal).
$\Theta$ is a threshold function defined as follows:

\begin{equation}\label{eq:tk:theta}
\Theta(n_1,n_2) = 
\left\{\def\arraystretch{1.2}
\begin{array}{ll}
0~:~if~sim(n_1, n_2) < \theta \\
1~:~if~sim(n_1, n_2) \geq \theta \\
\end{array}
\right.
\end{equation}

where $sim$ is a vector similarity function applied to the word embedding vectors of the input node pair $n_1$ and $n_2$, and $\theta$ is the similarity threshold above which the two nodes are considered equal for the kernel computation.\footnote{In fact, $sim$ can be any similarity measure between two words, not only a vector similarity between their embeddings vectors. However, the range of threshold values must be consistent with the output range of the similarity function.}

Similar to the string-match-based any-gram kernel described in Section~\ref{sec:anygram}, we first design a new method which eliminates the dependence on tree kernels and works directly with textual input.
The $\Delta$ for the new kernel function is defined as:

\begin{equation}\label{eq:agk:delta-west}
\Delta(S_1^i,S_2^j) = 
\left\{\def\arraystretch{1.5}
\begin{array}{ll}
0~:~if~\Theta(S_1^i, S_2^j) = 0~~
\vspace{.5em}
\\
\lambda \times (1+\Delta(S_1^{i+1},S_2^{j+1}))~:\\~if~\Theta(S_1^i, S_2^j) = 1
\end{array}
\right.
\end{equation}

The $\Theta$ is defined in the same way as in Equation \ref{eq:tk:theta}, with nodes replaced with tokens.
Algorithm \ref{algo:agk-west} shows the implementation of the new formulation.
The only change compared to Algorithm \ref{algo:agk-sm} is that the string match in the $if$ statement in the inner loop is replaced with a function $theta$ which is equivalent to $\Theta$ in Equation \ref{eq:tk:theta}.
As stated earlier, $\textproc{sim}$ can be any similarity measure between two inputs.

\begin{algorithm*}[t]
	\caption{: Any-gram kernel algorithm based on word embedding similarity threshold (WEST)}\label{algo:agk-west}
	\begin{algorithmic}
		\Function{anygram\_kernel\_west}{$S_1, S_2, theta$}
		\Comment{Inputs: tokenized sentence 1 and sentence 2, threshold}
		\State $kernel \gets 0$
		\State $delta[1, ..., len(S_1) + 1][1, ..., len(S_2) + 1] \gets 0$
		\Comment{Initializing $\Delta$ table}
		\State $L_1 \gets len(S_1)$
		\State $L_2 \gets len(S_2)$
		\For{$i \in \{L_1, ..., 1\}$}
		\Comment{Looping through sentence 1}
		\For{$j \in \{1, ..., L_2\}$}
		\Comment{Looping through sentence 2}
		\If{$\textproc{theta}(S_1[i], S_2[j], theta) = 1$}
		\Comment{Compute $\Delta$ only if tokens are similar}
		\State $delta[i][j] \gets \lambda \times (1 + delta[i+1][j+1])$
		\State $kernel \gets kernel + delta[i][j]$
		\EndIf
		\EndFor
		\EndFor
		\State \Return $kernel$
		\EndFunction
		\vspace*{1em}
		\Function{theta}{$S_1[i], S_2[j], theta$}
		\Comment{Inputs: tokens i of sentence 1 and j of sentence 2, threshold}
		\If{$\textproc{sim}(S_1[i], S_2[j]) \geq theta$}~~\Return 1
		\Else~~\Return 0
		\EndIf
		\EndFunction
	\end{algorithmic}
\end{algorithm*}

The threshold-based approach, however, has a few disadvantages. 
Firstly, it imposes an additional hyper-parameter to be tuned. 
Secondly, there can be word pairs where their similarity does not reach the set threshold, but they actually are perfectly similar in the context.
For example, the cosine similarity of \textit{superb} and \textit{brilliant} in our experiments is 0.72.
However, if the optimum threshold is found to be 0.80, this similarity will be ruled out.
From another perspective, such word pairs with similarity score only slightly lower than the threshold will have the same effect on the kernel value as those which are significantly less or not similar at all.

We modify the any-gram kernel function in a way which alleviates these problems.
Instead of a value of 1 for similar and 0 for different vectors based on the threshold value, we plug the similarity value itself into the kernel computation.
This eliminates the need for threshold tuning, takes into account the below-threshold similarities and factors in the extent of the similarity.
The formal definition of the new kernel function is as follows:
\begin{equation}\label{eq:agk:delta-wess}
\Delta(S_1^i,S_2^j) = 
\lambda \times (sim(S_1^i,S_2^j)+\Delta(S_1^{i+1},S_2^{j+1})
\end{equation}

It can be seen that the value $1$ in Equation \ref{eq:agk:delta} and Equation \ref{eq:agk:delta-west}, which accounted for the fact that the two tokens were the same or similar is replaced with the actual similarity score of the tokens, accounting for the extent to which they are similar. 
The implementation is presented in Algorithm \ref{algo:agk-wess}.

\begin{algorithm*}[t]
	\caption{: Any-gram kernel algorithm based on word embedding similarity score (WESS)}\label{algo:agk-wess}
	\begin{algorithmic}
		\Function{anygram\_kernel\_wess}{$S_1, S_2$}
		\Comment{Inputs: tokenized sentence 1 and sentence 2}
		\State $kernel \gets 0$
		\State $delta[1, ..., len(S_1) + 1][1, ..., len(S_2) + 1] \gets 0$
		\Comment{Initializing $\Delta$ table}
		\State $L_1 \gets len(S_1)$
		\State $L_2 \gets len(S_2)$
		\For{$i \in \{L_1, ..., 1\}$}
		\Comment{Looping through sentence 1}
		\For{$j \in \{1, ..., L_2\}$}
		\Comment{Looping through sentence 2}
		\State $delta[i][j] \gets \lambda \times (\textproc{sim}(S_1[i], S_2[j]) + delta[i+1][j+1])$
		\State $kernel \gets kernel + delta[i][j]$
		\EndFor
		\EndFor
		\State \Return $kernel$
		\EndFunction
	\end{algorithmic}
\end{algorithm*}

\section{Experiments}\label{sec:expr}

In this section, we carry out sentiment polarity classification experiments using the methods described in the previous section.
First, the data sets used for the experiments are described.
This is followed by the experimental settings and the results.

\subsection{Data}\label{sec:expr:data}

We use two slightly different problem sets: aspect-based sentiment classification and sentence-level sentiment classification.
For the former, we use the two data sets provided for Task 4 of SemEval 2014 \cite{Pontiki-EtAl-SemEval-2014} and for the latter we use the Stanford Sentiment Treebank \cite{Socher-EtAl-EMNLP-2013}.

\paragraph{Aspect-based sentiment analysis}

This task is the problem of sentiment analysis at a finer level of granularity than the sentence level.
Aspect terms are in fact targets towards which a sentiment is expressed in the text.
For instance, in the example in Figure \ref{fig:ngtk}, \textit{service} is an aspect term towards which the reviewer has expressed a positive sentiment.
Subtask 2 of Task 4 of SemEval2014 (called SE14 hereafter) dealt with predicting the polarity of the sentiment expressed towards the aspect terms (already given) in consumer reviews.
The polarity is categorized into four classes: \textit{positive, negative, neutral, conflict}.
The organizers released two data sets, one from laptop and one from restaurant reviews.
We use both of these data sets, which were also used in the previous work \cite{Kaljahi-Foster-PEOPLES-2016}.

Each data set comes in two subsets, one for training and one for test.
For development purposes, we use a subset of the training set which was also used in the original any-gram kernel paper \cite{Kaljahi-Foster-PEOPLES-2016} to have comparable results.
Table \ref{tab:data:se14} summarizes the statistics about the data sets.

\begin{table}[t]
	\begin{center}
		\begin{tabular}{l|rr|rr}
			\hline
			&	\multicolumn{2}{c|}{Laptop}
			&	\multicolumn{2}{c}{Restaurant}
			\\
			&	Train	& Test		&	Train	& Test
			\\\hline
			\# sentences	&	3045	&	800		&	3041	&	800
			\\
			\# aspect terms	&	2358	&	654		&	3693	&	1134
			\\
			~~\% positive	&	42\%	&	52\%	&	59\%	&	65\%	
			\\
			~~\% negative 	&	37\%	&	20\%	&	22\%	&	17\%	
			\\
			~~\% neutral 	&	19\%	&	26\%	&	17\%	&	17\%	
			\\
			~~\% conflict	&	2\%		&	2\%		&	2\%		&	 1\%	
			\\\hline
		\end{tabular}
	\end{center}
	\caption{Number of sentences, aspect terms and their polarity distributions in the SE14 data sets; number of aspect terms is in fact the number of instances used during training and test.}
	\label{tab:data:se14}
\end{table}

\paragraph{Sentence-level sentiment analysis}

For the task of sentence-level polarity classification, we use the Stanford Sentiment Treebank (called SSTB hereafter).
This data set is based on the movie review data set introduced by \newcite{Pang-Lee-ACL-2005} and was originally developed to analyze the compositional effect of language by labelling all constituents in parse trees with a sentiment polarity label from five categories: \textit{very positive, positive, neutral, negative, very negative}.
The label assigned to the root of the parse trees is viewed as the sentiment of the overall sentence.

The data set is released with training, development and test splits.
The statistics about this data set is presented in Table \ref{tab:data:sstb}

\begin{table}[t]
	\begin{center}
		\begin{tabular}{l|rrr}
			\hline
			\\
			&	Train	&	Dev		&	Test
			\\\hline
			\# sentences		&	8544	&	1101	&	2210
			\\
			~~\% very positive	&	15\%	&	15\%	&	18\%
			\\
			~~\% positive		&	27\%	&	25\%	&	23\%
			\\
			~~\% neutral		&	19\%	&	21\%	&	18\%
			\\
			~~\% negative		&	26\%	&	26\%	&	29\%
			\\
			~~\% very negative	&	13\%	&	13\%	&	12\%
			\\\hline
		\end{tabular}
	\end{center}
	\caption{Number of sentences and their polarity distributions in the SSTB data sets}
	\label{tab:data:sstb}
\end{table}

\subsection{Experimental Settings}\label{sec:expr:settings}

We use the any-gram kernel with \texttt{SVC}, a support vector machine implementation in the scikit-learn library\footnote{\url{http://scikit-learn.org/stable/modules/svm.html}}, which uses the \textit{one-versus-one} (OVO) method for multi-class classification, by default.
SVM's error/margin trade-off parameter ($C$) is tuned on the development sets of both data sets.
We use cosine similarity for word embedding similarities and the \textit{GloVe} \cite{Glove-EMNLP-2014} \textit{Common Crawl} (1.9M vocabulary) word embeddings with a dimensionality of 300.\footnote{We also experimented with pre-built word embeddings on Google News with the same dimensionality and also sentiment-specific word embeddings released by \newcite{Tang-EtAl-ACl-2014}. The GloVe Common Crawl vectors, however, performed better.}. 

It should be noted here that the aspect-based sentiment classification requires the aspect terms to be input to the learning algorithm.
The special tree representation used for the original any-gram kernel approach allows this sort of information to be incorporated in the input tree.
In the previous work \cite{Kaljahi-Foster-PEOPLES-2016}, we achieved this by inserting a node labelled $AT$ under those representing the aspect term tokens.
For instance, in the case of example in Figure \ref{fig:ngtk}, $AT$ replaces the node $X$ under \textit{service}, to indicate that \textit{service} is the aspect term.
With the new implementation, however, a different approach is needed to accomplish this.
In the case of the string-match-based any-grams, this can be achieved by simply attaching a suffix to the aspect term tokens.
For instance, the sentence in Figure \ref{fig:ngtk} with \textit{service} as aspect term is represented as \textit{Good, fast service\_AT.}

In the case of similarity-based any-grams, this suffix technique will however create unknown tokens for the pre-trained word embeddings.
To address this issue, we add an aspect term identifier to the word embedding vectors: if a token is an aspect term in the sentence, a 1 is added to the end of its word vector, and a 0 otherwise.
Therefore, the dimension of the word vectors will increase from $d$ to $d+1$.
This approach can be used to incorporate any kind of auxiliary input into the kernel function.

To compare the any-grams to state-of-the-art neural learning approaches, we build two LSTM  \cite{Hochreiter-Schmidhuber-NC-1997} models and apply them to the same data sets.
One model is unidirectional, where information only flows forward in time (i.e., from the previous word to the next in a sentence) and the other is bidirectional, where it flows in both directions to model the dependence of previous words on the next words.
The input to the models are the same pre-trained word embeddings used by the any-gram models.
A dropout layer is added before the output layer for regularization.
The hyper-parameters, summarized in Table \ref{tab:dnn:hyperparams}, are tuned on the same development sets as the any-gram models.
We used the \textit{Adam} \cite{Kingma-Ba-ARXIV-2014} optimization algorithm, a softmax function at the output layer, and cross-entropy as the loss function.
The models are built using \textit{Keras}\footnote{\url{https://keras.io/}} with a \textit{TensorFlow}\footnote{\url{https://www.tensorflow.org/}} backend.

\subsection{Results}
\begin{table*}[t]
	\begin{center}
		\begin{tabular}{l|ll|ll|ll}
			\hline
			&	\multicolumn{2}{c|}{Laptop}
			&	\multicolumn{2}{c|}{Restaurant}
			&	\multicolumn{2}{c}{SSTB}
			\\
			&	Dev			&	Test		&	Dev			& 	Test		&	Dev			& Test
			\\\hline
			\texttt{NGTK$_{sm}$}
			&	65.04		&	60.24		&	67.26		&	70.72		&	-			&	-
			\\
			\texttt{NGTK$_{west}$}
			&	64.78		&	62.08		&	68.40		&	70.99		&	-			&	-
			\\\hline
			\texttt{AGK$_{sm}$}
			&	62.47		&	63.15		&	68.88		&	73.10		&	38.96		&	40.95
			\\
			\texttt{AGK$_{west}$}
			&	62.98		&	62.99		&	68.88		&	73.36		&	38.33		&	40.50
			\\
			\texttt{AGK$_{wess}$}
			&	65.04		&	65.29		&	70.34		&	74.60		&	43.96		&	42.99
			\\\hline
			\texttt{UniLSTM}
			&	67.10		&	64.37		&	71.96		&	74.43		&	39.60		&	41.13
			\\
			\texttt{BiLSTM}
			&	66.58		&	65.60		&	71.15		&	75.66		&	42.69		&	43.57
			\\\hline
		\end{tabular}
	\end{center}
	\caption{Accuracy of tree-kernel-based any-grams \newcite{Kaljahi-Foster-PEOPLES-2016} (\texttt{NGT} prefix), any-gram kernels introduced here (\texttt{AGK} prefix) and LSTM models (\texttt{LSTM} suffix) for various settings, on SE14 and SSTB data sets} 
	\label{tab:results}
\end{table*}
\begin{table*}[t]
	\newcolumntype{C}{>{\centering\arraybackslash}p{1.3em}}
	\begin{center}
		\begin{tabular}{l|CCCCC|CCCCC|CCCCC}
			\hline
			& \multicolumn{5}{c|}{Laptop}
			& \multicolumn{5}{c|}{Restaurant}
			& \multicolumn{5}{c}{SSTB}
			\\\hline
			& \rotatebox[origin=l]{90}{batch size}
			& \rotatebox[origin=l]{90}{\#hidden layer}
			& \rotatebox[origin=l]{90}{hidden layer size~~~}
			& \rotatebox[origin=l]{90}{learning rate}
			& \rotatebox[origin=l]{90}{dropout rate}
			& \rotatebox[origin=l]{90}{batch size}
			& \rotatebox[origin=l]{90}{\#hidden layer}
			& \rotatebox[origin=l]{90}{hidden layer size~~~}
			& \rotatebox[origin=l]{90}{learning rate}
			& \rotatebox[origin=l]{90}{dropout rate}
			& \rotatebox[origin=l]{90}{batch size}
			& \rotatebox[origin=l]{90}{\#hidden layer}
			& \rotatebox[origin=l]{90}{hidden layer size~~~}
			& \rotatebox[origin=l]{90}{learning rate}
			& \rotatebox[origin=l]{90}{dropout rate}
			\\\hline
			\texttt{UniLSTM}
			& 40	& 1		& 50	& 0.001		& 0.3
			& 40	& 3		& 150	& 0.001		& 0.3
			& 40	& 1		& 100	& 0.01		& 0.3
			\\
			\texttt{BiLSTM}
			& 20	& 1		& 100	& 0.001		& 0
			& 20	& 1		& 100	& 0.001		& 0.3
			& 30	& 2		& 100	& 0.001		& 0.5
			\\\hline
		\end{tabular}
	\end{center}
	\caption{Hyper-parameters of the deep neural net models tuned on development sets (Up to 3 hidden layers were tried. The activation function for all settings is $tanh$.)} 
	\label{tab:dnn:hyperparams}
\end{table*}

Table \ref{tab:results} illustrates the performance of all the models built in these experiments as well as those we reported for the tree-kernel-based any-grams in \cite{Kaljahi-Foster-PEOPLES-2016}.
Systems in the first and second rows of the table, with their names starting with \texttt{NGTK}, are the ones based on tree-kernel-based any-grams and those in the middle section of the table, with their names starting with \texttt{AGK}, are the ones built using the new any-gram kernels.
The last two rows in the table are for the LSTM models.
The $sm$ suffix stands for string match, the $west$ for word embedding similarity threshold and the $wess$ for word embedding similarity score.

First, we compare the effect of using the new implementation of any-gram kernels, independent of tree kernels. i.e. \texttt{NGTK$_{sm}$} with \texttt{AGK$_{sm}$} and \texttt{NGTK$_{west}$} with \texttt{AGK$_{west}$}.
We can see that the new method outperforms the original one for the majority of the data sets, with improvements of up to 2.9 percent.
However, the accuracy on the laptop development set has been degraded by 2.5 percent.
Although the primary goal behind designing the new implementation for the any-gram kernel was the efficiency, it turns out that it leads to a better performance as well.
It should also be noted that, similar to the original any-grams, the threshold-based use of word embedding similarity is not effective with the new implementation either.

Next, we verify the effectiveness of our proposed method for using word embedding similarities with any-gram kernels by comparing \texttt{AGK$_{sm}$} and \texttt{AGK$_{west}$} with \texttt{AGK$_{wess}$}.
As can be seen, the similarity-score-based system outperforms both systems on all 6 data sets, the improvements ranging from 1.2 to 5.6 percent in accuracy.
This systems also recovers the afore-mentioned loss on the laptop development set compared to the original any-grams.

Finally, the best any-gram kernel, \texttt{AGK$_{wess}$}, achieves competitive results with the LSTM models.
It should however be noted that the LSTM models have a large set of hyper-parameters which require extensive tuning to obtain the optimum performance, whereas the similarity-score-based any-grams have only the $C$ parameter of the SVM to be optimized.

\section{Conclusion}\label{sec:conclusion}

We introduced a new formulation of any-gram kernels which 1) is independent of tree kernels and does not require the input to be represented in the special binary tree form, and 2) makes better use of word-vector-based comparison to replace mere string match.
The new method outperformed the original implementation of any-gram kernels and achieved comparable results to LSTM models across three widely used sentiment analysis data sets.
We presented efficient implementations for the variations of the new any-gram kernel.

So far we have applied any-gram kernels to sentiment polarity classification but it could be applied to other text analysis problems including both classification and regression.
It is also worth noting that our experiments only used the n-grams.
Combining the any-gram kernels with other features could be beneficial.
One way to achieve this combination is via auxiliary input to the any-gram kernels which we described in the context of marking the aspect term in a sentence.
Another way would be to combine the any-gram kernel with other more traditional kernels (e.g. RBF) applied to hand-crafted features.

\section*{Acknowledgments}

This research is supported by Science Foundation Ireland in the ADAPT Centre (Grant 13/RC/2106) (\url{www.adaptcentre.ie}) at Dublin City University.

\bibliographystyle{acl2014}
\bibliography{arXiv2017}

\end{document}